\def\year{2022}\relax
\documentclass[letterpaper]{article} 
\usepackage{aaai22}  
\usepackage{times}  
\usepackage{helvet}  
\usepackage{courier}  
\usepackage[hyphens]{url}  
\usepackage{graphicx} 
\usepackage{natbib}  
\usepackage{caption} 
\DeclareCaptionStyle{ruled}{labelfont=normalfont,labelsep=colon,strut=off} 
\usepackage{algorithm}
\usepackage{algorithmic}
\usepackage{multirow}
\usepackage{color}
\usepackage{makecell}
\usepackage{booktabs}
\usepackage{amsmath}
\usepackage{amssymb}
\usepackage{bbding}
\usepackage{newfloat}
\usepackage{listings}
\usepackage{colortbl}
\floatstyle{ruled}
\newfloat{listing}{tb}{lst}{}
\floatname{listing}{Listing}
\def\@onedot{\ifx\@let@token.\else.\null\fi\xspace}

\definecolor{bb}{rgb}{0.0, 0.0, 0.5}
\usepackage[colorlinks=true,linkcolor=red,citecolor=bb,urlcolor=red]{hyperref}

\definecolor{Gray}{gray}{0.9}

\newcommand{\ch}[1]{{\color{black}#1}}

\title{Disentangled Feature Representation for Few-shot Image Classification}

\author{
    Hao Cheng\textsuperscript{\rm 1},
    Yufei Wang\textsuperscript{\rm 1},
    Haoliang Li\textsuperscript{\rm 2},
    Alex C. Kot\textsuperscript{\rm 1},
    Bihan Wen\textsuperscript{\rm 1}\thanks{Bihan Wen is the corresponding author.}
    \\
}
\affiliations {
    \textsuperscript{\rm 1} School of Electrical and Electronic Engineering, Nanyang Technological University, Singapore \\
    \textsuperscript{\rm 2} Department of Electrical Engineering, City University of Hong Kong, China \\
     \{hao006, yufei001, eackot, bihan.wen\}@ntu.edu.sg, \quad haoliang.li@cityu.edu.hk
}

\begin{document}

\maketitle

\begin{abstract}

Learning the generalizable feature representation is critical for few-shot image classification.
While recent works exploited task-specific feature embedding using meta-tasks for few-shot learning, they are limited in many challenging tasks as being distracted by the excursive features such as the background, domain and style of the image samples. 
In this work, we propose a novel Disentangled Feature Representation framework, dubbed DFR, for few-shot learning applications.
DFR can adaptively decouple the discriminative features that are modeled by the classification branch, from the class-irrelevant component of the variation branch. 
In general, most of the popular deep few-shot learning methods can be plugged in as the classification branch, thus DFR can boost their performance on various few-shot tasks.
Furthermore, we propose a novel FS-DomainNet dataset based on DomainNet, for benchmarking the few-shot domain generalization tasks.
We conducted extensive experiments to evaluate the proposed DFR on general and fine-grained few-shot classification, as well as few-shot domain generalization, using the corresponding four benchmarks, i.e., mini-ImageNet, tiered-ImageNet, CUB, as well as the proposed FS-DomainNet.
Thanks to the effective feature disentangling, the DFR-based few-shot classifiers achieved the state-of-the-art results on all datasets.


\end{abstract}

\section{Introduction}

While deep neural networks achieved superior results on image classification via supervised learning from large-scale datasets, it is challenging to classify a query sample using only few labelled data, which is known as \textit{few-shot classification}~\cite{fei2006one}.
How to learn the discriminative feature representation that can be generalized from the training set to new classes in testing is critical for few-shot tasks.
Popular few-shot methods applied \textbf{\emph{meta-learning}}~\cite{vinyals2016matching} by episodic training from a large amount of simulated meta-tasks, to obtain a task-specific feature embedding associated with a distance metric (e.g., cosine or Euclidean distance) for classification.

\begin{figure}[t]
\centering
\includegraphics[width=1.0\columnwidth]{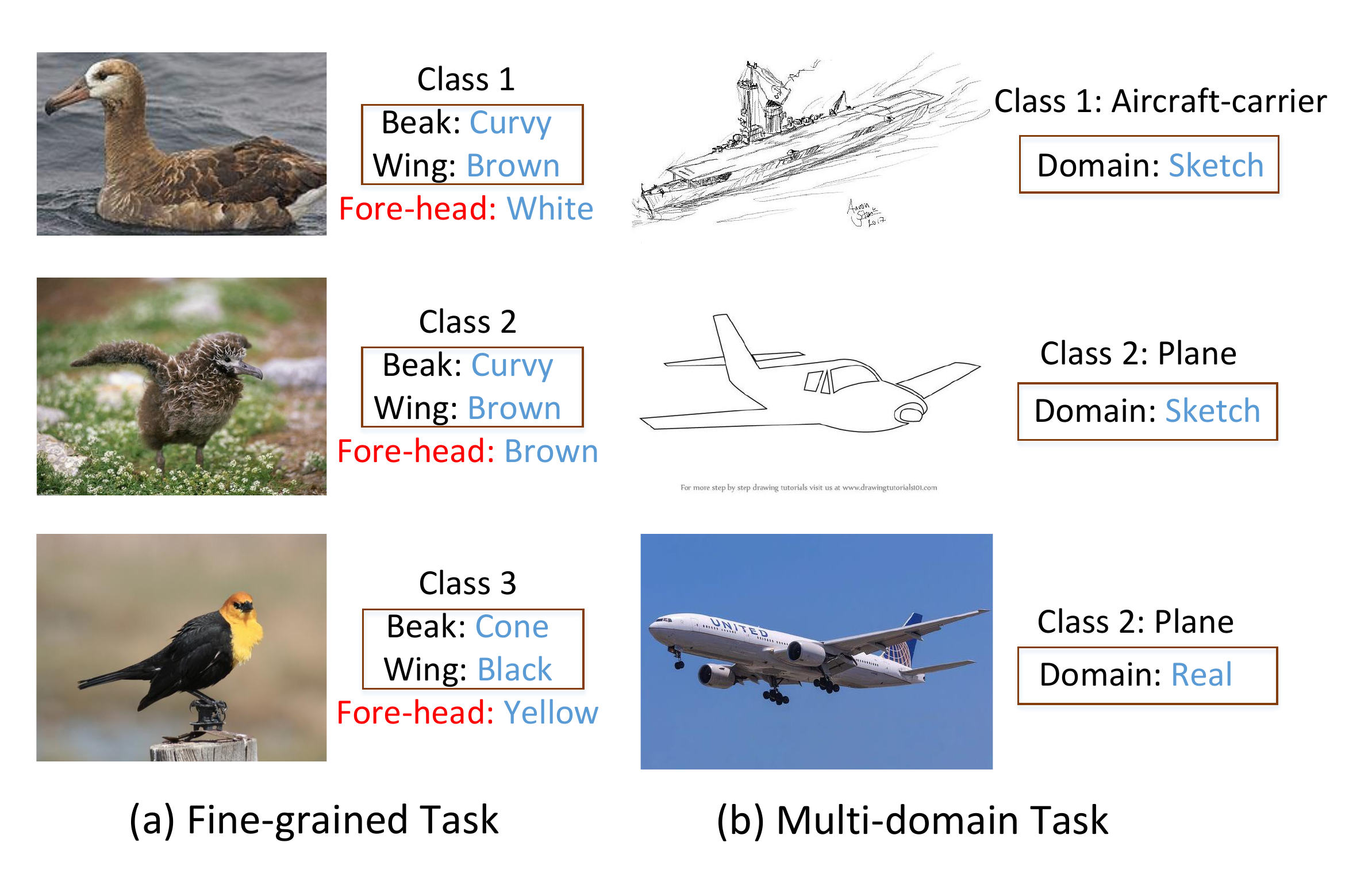}
\caption{Excursive features (highlighted in boxes) that distract few-shot classification for fine-grained and multi-domain tasks.}
\label{fig:intro}
\end{figure}

In practice, many excursive features of image data, e.g., style, domain and background, are typically class-irrelevant. 
Figure~\ref{fig:intro} shows two such examples in fine-grained and multi-domain classification tasks, respectively, which are challenging for few-shot learning:
(1) Only the subtle traits are critical to characterize and differentiate the objects of fine-grained classes; 
(2) The style and domain information dominates the image visual presence, but they are in fact the excursive and class-irrelevant features.
As the subtle traits vary in different simulated meta-tasks, they can hardly be preserved by the learned embedding.
On the contrary, the excursive features usually distract the feature embedding~\cite{tokmakov2019learning,Zhang_2020_CVPR}, leading to the degraded few-shot classification results.
To rectify such limitations, most recent few-shot methods attempted to suppress excursive features or propose proper metrics, e.g., LCR~\cite{tokmakov2019learning}, DeepEMD~\cite{Zhang_2020_CVPR}, FEAT~\cite{ye2020fewshot} and CNL~\cite{zhao2021looking}.
However, none of the existing methods explicitly extract the class-specific representation from the excursive image features. 

In this paper, we present a novel approach to incorporate deep disentangling for few-shot image classification. Such approach can selectively extract the subtle traits for each task, while maintaining the model generalization.
First, we propose a novel Disentangled Feature Representation (DFR) framework which can be applied to most few-shot learning methods. 
DFR contains two branches: the \textit{classification branch} extracts the discriminative features of the image sample, while the \textit{variation branch} encodes the class-irrelevant information that complements the image representation.
A RelationNet~\cite{sung2018learning} is applied in the variation branch to measure the feature similarity of each sample pair.
A hybrid loss is applied for training DFR, including a reconstruction loss to ensure image information preservation, as well as the translation, discriminative and cross-entropy losses for class-specific feature disentangling.
At the inference stage, only the disentangled features from the classification branch are used for class prediction.
Second, we integrate the proposed DFR framework into representative baselines for few-shot classification, including the popular ProtoNet~\cite{snell2017prototypical} and the state-of-the-art DeepEMD~\cite{Zhang_2020_CVPR} and FEAT~\cite{ye2020fewshot}, to carefully investigate the behaviour of DFR with feature visualization and analysis.
Extensive experiments are conducted on a set of few-shot tasks, i.e., general image classification, fine-grained image classification, and domain generalization over four benchmarks to demonstrate the effectiveness of our DFR framework.

Our main contributions are summarized as follows:
\begin{itemize}
 \item We propose a novel disentangled feature representation (DFR) framework, which can be easily applied to most of the few-shot learning methods to extract class-specific features from excessive information.
 
 \item We propose a novel benchmark named FS-DomainNet based on DomainNet~\cite{peng2019moment} \ch{and fully study the few-shot domain generalization task with two evaluation settings.}
 
 \item We evaluate the DFR framework over four few-shot benchmarks, i.e., mini-ImageNet, tiered-ImageNet, CUB-200-2011, and the proposed FS-DomainNet dataset.
 Results show that incorporating DFR into existing few-shot algorithms, including both baseline and state-of-the-art methods, can generate consistent gains for multi few-shot classification tasks under both 5-way 1-shot and 5-way 5-shot settings.
 
 

\end{itemize}

\begin{figure*}[t]
\centering
\includegraphics[trim=0cm 7cm 0cm 1.0cm,  width=1\textwidth]{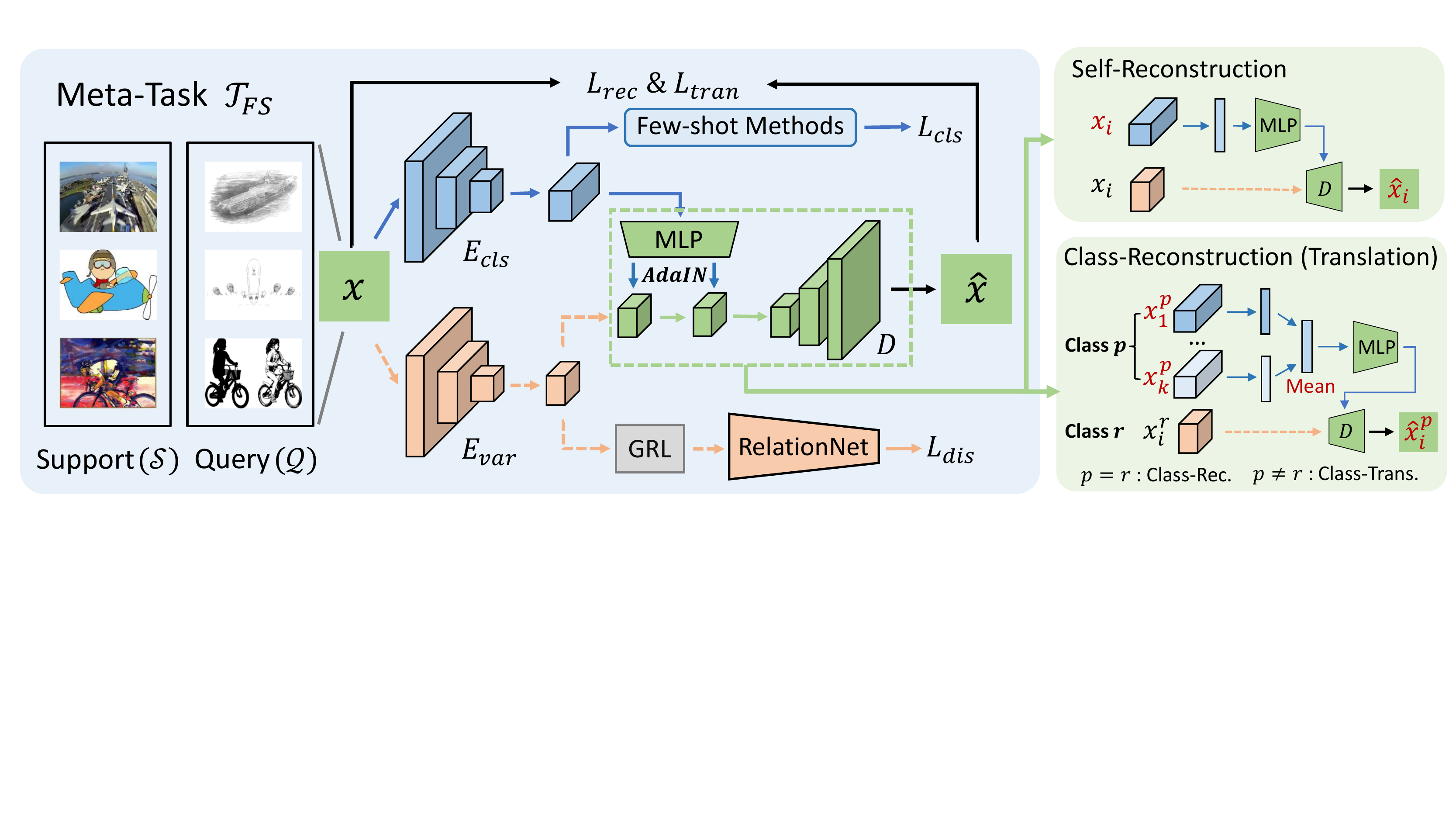}
\caption{DFR for few-shot image classification: Given a few-shot meta-task $\mathcal{T}_{FS}$ with support ($\mathcal{S}$) and query ($\mathcal{Q}$) image sets, the encoders ($E_{cls}$ and $E_{var}$) of the classification and variation branches extract the class-specific and class-irrelevant features, respectively. $E_{cls}$ is a classic backbone used in the few-shot methods (e.g., ResNet-12 in this work) which follows the blue stream. The output of $E_{cls}$ is used for few-shot classification, and the output of $E_{var}$ is fed to the RelationNet with a gradient reverse layer (GRL) to remove any class-related information. An MLP block extracts the class information from the classification branch to guide the image reconstruction and translation. Specifically, the Decoder D can achieve self-reconstruction, class-reconstruction, or class-translation according to the different inputs of MLP.
}
\label{fig:frame}
\end{figure*}

\section{Related Work}

\subsection{Few-shot Learning}
According to the meta-learning framework~\cite{vinyals2016matching}, there are mainly three types of few-shot learning methods.
Firstly, the gradient-based methods utilize a good model initialization~\cite{finn2017model,nichol2018first} or optimization strategy~\cite{ravi2016optimization,rusu2018meta,lee2019meta,Liu2020E3BM} to quickly adapt to novel tasks.
Secondly, the data augmentation-based methods focus on generating~\cite{gidaris2019generating,li2020adversarial} or gathering augmented data~\cite{hariharan2017low,wang2018low,yang2021free} to enable classification from limited samples.
In this work, we focus on the third type, namely the metric learning-based methods, i.e., to learn the discriminative feature embedding for distinguishing different image classes.
For example, ProtoNet~\cite{snell2017prototypical} considered the class-mean representation as the prototype of each class and applied the Euclidean distance metric for classification.
LCR~\cite{tokmakov2019learning} applied the subspace-based embedding for each class and DeepEMD~\cite{Zhang_2020_CVPR} adopted the earth mover's distance as the metric function to compare the similarity between two feature maps in a structured way.
FEAT~\cite{ye2020fewshot} defined four kinds of set-to-set transformation including self-attention transformer~\cite{jaderberg2015spatial} to learn a task-specific feature embedding for few-shot learning.
Based on prior knowledge, COMET~\cite{cao2021concept} mapped some high-level visual concepts into a semi-structured metric space and then learned an ensemble classifier by combining the outputs of independent concept learners.
Tang et al.~\cite{tang2020revisiting} also uses a semi-structured feature space based on independent prior knowledge concepts to do pose normalization for fine-grained tasks.
Our work does not intend to propose new metrics, but focuses on extracting the class-specific features from the variations distracting the metric learning, thus to improve few-shot classification. 

\subsection{Disentangled Feature Representations}
Disentangled feature representation aims to learn an interpretable representation
for image variants, which has been widely studied in tasks such as face generation~\cite{chen2016infogan}, style translation~\cite{lee2020drit++,liu2019few}, image restoration~\cite{li2020learning}, 
video prediction~\cite{hsieh2018learning} and image classification~\cite{prabhudesai2021disentangling,li2021generalized}.
InfoGAN~\cite{chen2016infogan} applied an unsupervised method to learn interpretable and disentangled representations by maximized mutual information.
DRIT~\cite{lee2020drit++} embedded images into a content space and a domain-specific attribute space and applied a cycle consistency loss for style translation.
FDR~\cite{li2020learning} applied channel-wise feature disentanglement to reduce the interference between hybrid distortions for hybrid-distorted image restoration.
Li et al.~\cite{li2021generalized} proposed a disentangled-VAE to excavate category-distilling information from visual and semantic features for generalized zero-shot learning.

It is noteworthy that the very recent D3DP~\cite{prabhudesai2021disentangling} also adopted a feature disentangling scheme for few-shot detection and VQA, by dividing high-dimensional data (e.g., RGB-D) into individual objects and other attributes.
However, our DFR significantly differs from D3DP in the following aspects: 
DFR is a general-purpose feature extractor for image classification, while D3DP only disentangles individual object to tackle more specific object detection and VQA in 3D scenes.
Besides, our DFR works on real images from few-shot benchmarks while D3DP only works with synthetic scenes.
Moreover, the DFR framework is much simpler comparing to D3DP with fewer parameters and more efficient algorithms. Thus DFR can serve an enhanced feature extractor for classic backbones that are widely used in most of the existing few-shot methods.

\section{Proposed Method}

In this section, we start with a brief introduction to few-shot learning.
Then the proposed DFR framework is explained in detail, followed by the loss function of our model and why our DFR works well.

\subsection{Problem Definition}
Given a training image set with the base classes $\mathcal{C}_{train}$, few-shot image classification task aims to predict the novel classes $\mathcal{C}_{test}$ from the testing set, i.e., $\mathcal{C}_{train} \cap \mathcal{C}_{test} = \emptyset$.
Thus, the trained classifier from $\mathcal{C}_{train}$ needs to be generalized to $\mathcal{C}_{test}$ in the testing stage with only few labeled samples.
In this paper, we follow the meta-learning strategy (i.e., the $N$-way $K$-shot setting)~\cite{vinyals2016matching} to simulate meta-tasks in the training set that are similar to the few-shot setting at the testing stage, i.e., each meta-task $\mathcal{T}_{FS}$ contains a support set $\mathcal{S}$, and a query set $\mathcal{Q}$.
The support set $\mathcal{S}$ contains $N$ classes with $K$ labeled samples ($N$ and $K$ are both very small) and a query set $\mathcal{Q}$ with unlabeled query samples from $N$ classes is used to evaluate the performance. 

\subsection{Disentangled Feature Representation}
Figure~\ref{fig:frame} is an overview of the proposed DFR framework. 
With a few-shot task $\mathcal{T}_{FS}$ of support set $\mathcal{S}$ and query set $\mathcal{Q}$, the objective is to extract discriminative features for classification from the excursive information of each image $x_i$.
The proposed DFR consists of two branches with two encoders, i.e., $E_{cls}$ and $E_{var}$ for the classification and variation branches, respectively, and one decoder $D$, as well as a discriminator with a gradient reverse layer and a relation module.

\noindent\textbf{Classification Branch.}
In principle, any classic metric-based backbone for few-shot learning can be applied as $E_{cls}$ in this branch, to extract the class-specific features of each $x_i$.
In this work, the commonly-used ResNet-12 backbone is adopted as $E_{cls}$, and the classifier $f(\cdot)$ varies for different few-shot learning baselines being used (e.g., ProtoNet~\cite{snell2017prototypical}, DeepEMD~\cite{Zhang_2020_CVPR} and FEAT~\cite{ye2020fewshot} are applied in this work, with the corresponding models denoted as +DFR).
Therefore, the query sample $x_i^\mathcal{Q}$ can be classified based on the support samples $x^\mathcal{S}$ as
\begin{equation}
\hat{y}_{i} = f(E_{cls}(x_i^\mathcal{Q});\{E_{cls}(x^\mathcal{S}), y^\mathcal{S}\}).
\end{equation}

\noindent\textbf{Variation Branch.}
The role of the variation branch is to encode the class-irrelevant information of image samples, which consists of an encoder $E_{var}$ followed by a discriminator.
The feature map dimension (i.e., $h\times w$) of $E_{var}(x_i)$ is set to be higher than that of $E_{cls}(x_i)$, to contain more excursive image features.
Moreover, we apply instance normalization (IN) and adaptive instance normalization (AdaIN) instead of batch normalization to achieve information transfer for variation encoder $E_{var}$ and decoder $D$, respectively.
The discriminator is formed by a gradient reversal layer (GRL) and a RelationNet $r_\varphi$~\cite{sung2018learning} to measure the variation feature similarity between any two samples. 
To be specific, the GRL acts as an identity transform in forward pass, and it multiples the gradient from the subsequent level by a constant $\textnormal{-}\lambda$ during back-propagation.
In training, we construct positive and negative pairs from the meta-task by their real labels.
The relation module $r_\varphi$ outputs a score $s_{i}\in [0,1] $ indicating the probability that the pair $x_{i1}$ and $x_{i2}$ are from the same class as
\begin{equation}
s_{i}=r_\varphi \left(E_{var}(x_{i1}), E_{var}(x_{i2})\right).
\label{rn}
\end{equation}

\noindent\textbf{Decoder Module.}
To preserve the image information and achieve feature disentanglement, a decoder module with a MLP module and a decoder network $D$ combines the classification and variation branches for image reconstruction and translation.

To be specific, the output feature of the classification branch is fed to the MLP module $g$ to extract class-specific information $(\mu, \sigma)$ of each sample for scaling the feature of the variation branch in the follow-up decoder.
The decoder can reconstruct or translate the source image based on different sources of $(\mu, \sigma)$, shown in Figure~\ref{fig:frame}, as
\begin{equation}\label{eq:dec}
\hat{x}_i = D(E_{var}, g(X)),
\end{equation}
where $X$ can be the feature of the $i$-th sample itself for self-reconstruction, or the mean feature of class $y_i$ for class-reconstruction or another class $y_j$ with $j\neq i$ for class-translation.



\subsection{Loss Function}
The objective function consists of the discriminative loss $L_{dis}$, cross-entropy loss $L_{cls}$, reconstruction loss $L_{rec}$ and translation loss $L_{tran}$.

\noindent\textbf{Discriminative Loss.}
To remove the class-specific information in the variation branch, we incorporate the binary cross-entropy loss to optimize the variation feature maps based on the score of RelationNet as
\begin{equation}
L_{dis} = -\sum_{i=1}^{P} \left(l_i\cdot log(s_i)+(1-l_i)\cdot log(1-s_i)\right),
\end{equation}
where $P$ denotes the number of training pairs, $s_i$ is the relation score of the $i$-th pair which is calculated by (\ref{rn}), and $l_i=0$ or $1$ indicates the ground truth whether the $i$-th training pair is positive.
We minimize $L_{dis}$ in training, and apply GRL to reverse the gradient during back-propagation to achieve feature disentangling, i.e., to minimize the class-specific information captured by the variation feature.

\noindent\textbf{Cross-Entropy Loss.}
To preserve class-related features for few-shot classification, we minimize the cross-entropy loss $L_{cls}$ for the classification branch for query samples of all classes as
\begin{equation}
L_{cls}=-\sum_{i=1}^{Q} y_{i} \log P\left(\hat{y}_{i}=y_{i} \mid \mathcal{T}_{FS}\right),
\end{equation}
where $Q$ is the number of query samples in a meta-task $\mathcal{T}$, $y_i$ and $\hat{y_i}$ denote the true and predicted class label of each query sample $x_i$, respectively.

\noindent\textbf{Reconstruction and Translation Loss.}
To ensure that the disentangled classification and variation features can jointly restore the input image, an $\ell_1$-norm penalty for image reconstruction and a perceptual loss~\cite{johnson2016perceptual} are applied after decoding for self-reconstruction and class-reconstruction as
\begin{equation}
L_{rec}=\frac{1}{M} \sum_{i=1}^{M} \|x_i-\hat{x}_{i}\|_{1} + \frac{1}{M} \sum_{i=1}^{M} \|\phi(x_i)-\phi(\hat{x}_{i}^{c_i})\|_{1},
\end{equation}
where $M$ denotes the number of samples in a meta-task $\mathcal{T}_{FS}$.
$\hat{x}_{i}$ and $\hat{x}_{i}^{c_i}$ are the reconstructed images of $x_i$ based on the feature of the $i$-th sample itself and mean feature $c_i$ of class $y_{i}$ using (\ref{eq:dec}), respectively.

Moreover, the perceptual loss is also adapted to measure perceptual differences between the output image $\hat{x}_{i}^{c_j}$ and the support set of the $j$-th class for class-translation to achieve feature disentanglement as
\begin{equation}
L_{tran} = \frac{1}{N} \sum_{i=1}^{M} \sum_{l=1}^{K} \|\phi(x_{l}^{\mathcal{S}_j})-\phi(\hat{x}_{i}^{c_j})\|_{1}.
\end{equation}

The total loss for training DFR can be formulated as
\begin{equation}
L_{total}= \lambda_1\cdot L_{dis} + \lambda_2\cdot L_{rec} + \lambda_3\cdot L_{tran} + L_{cls},
\end{equation}
where $\lambda_1$, $\lambda_2$ and $\lambda_3$ denote the weights parameters of $L_{dis}$, $L_{rec}$ and $L_{tran}$ relative to $L_{cls}$, respectively.

\begin{figure}[!]
\centering
\includegraphics[trim=3.5cm 0cm 2.5cm 2cm,clip,width=1.0\columnwidth]{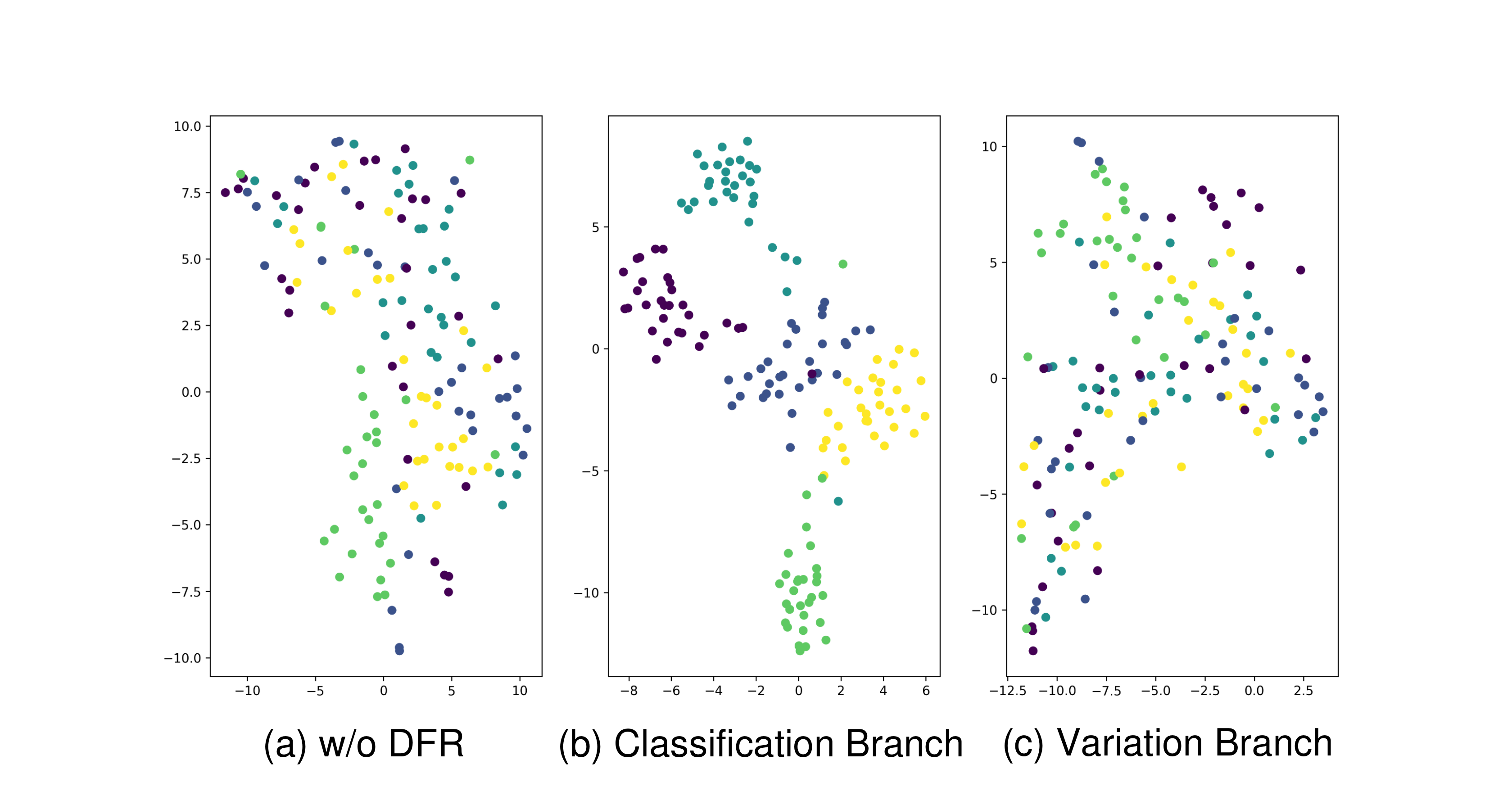}
\caption{The t-SNE visualization of the feature representations: (a) the learned features of ResNet-12 backbone for methods w/o DFR, (b) the output features of the classification branch, and (c) the output features of the variation branch.}
\label{fig:tsne}
\end{figure}

\begin{table*}[!t]
   \centering
   \setlength{\tabcolsep}{0.6em}
   \small
   \begin{tabular}{lllll}
     \toprule
     \multirow{2}{1cm}{\textbf{Method}} 
      & \multicolumn{2}{c}{mini-ImageNet}  & \multicolumn{2}{c}{tiered-ImageNet}  \\
     \cmidrule(r){2-3} \cmidrule(r){4-5}
       &  5-way 1-shot & 5-way 5-shot  & 5-way 1-shot & 5-way 5-shot \\
     \midrule
     \bf{TADAM}~\cite{oreshkin2018tadam}    & 58.50 \textit{$\pm$} 0.30 & 76.70 \textit{$\pm$} 0.30 & \quad\;\;\;\;\;\;- & \quad\;\;\;\;\;\;- \\
     \bf{AFHN}~\cite{li2020adversarial}    & 62.38 \textit{$\pm$} 0.72 & 78.16 \textit{$\pm$} 0.56 & \quad\;\;\;\;\;\;- & \quad\;\;\;\;\;\;- \\
     \bf{MetaOptNet}~\cite{lee2019meta}    & 62.64 \textit{$\pm$} 0.82 & 78.63 \textit{$\pm$} 0.46 & 65.99 \textit{$\pm$} 0.72 & 81.56 \textit{$\pm$} 0.53\\
     \bf{DSN}~\cite{simon2020adaptive}    & 62.64 \textit{$\pm$} 0.66 & 78.83 \textit{$\pm$} 0.45 & 66.22 \textit{$\pm$} 0.75 & 82.79 \textit{$\pm$} 0.48 \\  
     \bf{MatchNet}~\cite{vinyals2016matching}    & 63.08 \textit{$\pm$} 0.80 & 75.99 \textit{$\pm$} 0.60 & 68.50 \textit{$\pm$} 0.92 & 80.60 \textit{$\pm$} 0.71 \\
     \bf{E$^3$BM}~\cite{Liu2020E3BM}    & 63.80 \textit{$\pm$} 0.40 & 80.10 \textit{$\pm$} 0.30 & 71.20 \textit{$\pm$} 0.40 & 85.30 \textit{$\pm$} 0.30 \\
     \bf{CAN}~\cite{hou2019cross}    & 63.85 \textit{$\pm$} 0.48 & 79.44 \textit{$\pm$} 0.34 & 69.89 \textit{$\pm$} 0.51 & 84.23 \textit{$\pm$} 0.37 \\
     \bf{CTM}~\cite{li2019finding}   & 64.12 \textit{$\pm$} 0.82 & 80.51 \textit{$\pm$} 0.13 & 68.41 \textit{$\pm$} 0.39 & 84.28 \textit{$\pm$} 1.73\\
     \bf{P-Transfer}~\cite{shen2021partial}   & 64.21 \textit{$\pm$} 0.77 & 80.38 \textit{$\pm$} 0.59 & \quad\;\;\;\;\;\;- & \quad\;\;\;\;\;\;-\\
     \bf{RFS}~\cite{tian2020rethinking}   & 64.82 \textit{$\pm$} 0.60 & 82.14 \textit{$\pm$} 0.43 & 71.52 \textit{$\pm$} 0.69 & 86.03 \textit{$\pm$} 0.49\\
     \bf{ConstellationNet}~\cite{xu2020attentional}   & 64.89 \textit{$\pm$} 0.23 & 79.95 \textit{$\pm$} 0.17 & \quad\;\;\;\;\;\;- & \quad\;\;\;\;\;\;-\\
     \bf{FRN}~\cite{wertheimer2021few}   & 66.45 \textit{$\pm$} 0.19 & 82.83 \textit{$\pm$} 0.13 & 71.16 \textit{$\pm$} 0.22 & 86.01 \textit{$\pm$} 0.15\\
     \bf{infoPatch}~\cite{liu2021learning} & 67.67 \textit{$\pm$} 0.45 & 82.44 \textit{$\pm$} 0.31 & 71.51 \textit{$\pm$} 0.52 & 85.44 \textit{$\pm$} 0.35\\
     \midrule
     \bf{ProtoNet}~\cite{snell2017prototypical}  & 61.83 \textit{$\pm$} 0.20 & 79.86 \textit{$\pm$} 0.14 & 66.84 \textit{$\pm$} 0.23 & 84.54 \textit{$\pm$} 0.16 \\
     \rowcolor{Gray} \bf{ProtoNet\;+\;DFR}    & \bf{64.84 \textit{$\pm$} 0.20} & \bf{81.10 \textit{$\pm$} 0.14} & \bf{70.22 \textit{$\pm$} 0.23} & \bf{84.74 \textit{$\pm$} 0.16}\\
     \midrule
     \bf{DeepEMD}~\cite{Zhang_2020_CVPR}    & 64.93 \textit{$\pm$} 0.29 & 81.73 \textit{$\pm$} 0.57 & 70.47 \textit{$\pm$} 0.33 & 84.76 \textit{$\pm$} 0.61\\
     \rowcolor{Gray} \bf{DeepEMD\;+\;DFR}    & \bf{65.41 \textit{$\pm$} 0.28} & \bf{82.18 \textit{$\pm$} 0.55} & \bf{71.56 \textit{$\pm$} 0.31} & \bf{86.23 \textit{$\pm$} 0.58} \\
     \midrule
     \bf{FEAT}~\cite{ye2020fewshot}    & 66.52 \textit{$\pm$} 0.20 & 81.46 \textit{$\pm$} 0.14 & 70.30 \textit{$\pm$} 0.23 & 84.55 \textit{$\pm$} 0.16\\
     \rowcolor{Gray} \bf{FEAT\;+\;DFR}    & \bf{67.74 \textit{$\pm$} 0.86} & \bf{82.49 \textit{$\pm$} 0.57} & \bf{71.31 \textit{$\pm$} 0.93} & \bf{85.12 \textit{$\pm$} 0.64}\\ 
     \bottomrule
   \end{tabular}
    \caption{Few-shot classification accuracy ($\%$) averaged over mini-ImageNet and tiered-ImageNet with the ResNet backbone.} 
     \label{res1}
 \end{table*}

\subsection{Why It Works}
DFR framework aims to extract only class-related information for classification. Different from other attempts towards more adaptive embedding using attention mechanism~\cite{hou2019cross,li2019finding,ye2020fewshot}, our classification and variation branches play as the adversaries by minimizing $L_{cls}$ and $L_{dis}$ simultaneously. 
In practice, the classification and variation features of an image are always complementary, thus the image reconstruction quality is enforced after fusion by minimizing $L_{rec}$.
It is essential to preserve the image representation in DFR for few-shot classification: 
as the class-specific features can be task-varying thus hard to be generalized, any information loss throughout the inter-state flow may potentially limit the model performance. 
Such design is contrast to the classic feature embedding for few-shot learning, in which image features are always projected onto the lower-dimensional manifolds~\cite{simon2020adaptive}.
Our $E_{cls}$ feature has much lower dimension comparing to the $E_{var}$ feature, as the class-irrelevant information (e.g., image style and background) are typically excessive. To this end, a more restrictive classification feature will significantly reduce the model bias, thus to enhance its generalizability in few-shot tasks.

We visualize the feature distributions w/o and w/ DFR using t-SNE~\cite{van2008visualizing} to verify our intuition.
Figure~\ref{fig:tsne} (a) shows that the learned features extracted from the ResNet-12 backbone are less discriminative without using the DFR framework. 
While when applying the DFR framework, the classification branch clusters in Figure~\ref{fig:tsne} (b) are more separable from each other, and the output features of the variation branch in Figure~\ref{fig:tsne} (c) contains more class-irrelevant information that meets our expectations.



\section{Experiment}

We conduct extensive experiments on two few-shot benchmarks, i.e., Mini-ImageNet and Tiered-ImageNet on general few-shot classification tasks to evaluate the performance of our proposed DFR framework.
After that, we introduce a novel FS-DomainNet dataset with the proposed two evaluation settings for benchmarking few-shot domain generalization task (FS-DG).
Moreover, we evaluate the performance of DFR on CUB-200-2011 benchmark on fine-grained few-shot classification task.~\footnote{The code of the proposed DFR model and FS-DomainNet dataset will be available on https://github.com/chengcv/DFRFS.}.

\subsection{Implementation Details}

We use ResNet-12 network~\cite{lee2019meta} as our backbone $E_{cls}$ for classification branch and set the number of channels as $[64, 160, 320, 640]$, which are similar to the competing methods.
The encoder $E_{var}$ consists of four convolutional blocks and two residual blocks.
The decoder contains a two-layer MLP block and a decoder network $D$ with residual blocks and upscale convolutional blocks.
The I/O channel numbers of variation encoder and decoder are all set to $128$.
The level ratio $\lambda$ of GRL layer is set to $1$. 

Data augmentation including resizing, random cropping, color jitter and random flipping following~\cite{ye2020fewshot} are applied for all methods in training.
Our models are all trained using SGD optimizer, with the weight decay as $5e{-4}$, and the momentum as $0.9$.

We conduct experiments under both $5$-way $1$-shot and $5$-way $5$-shot settings with $15$ query images each class, i.e., $5\times(1\,and\,5)+5\times15$ samples for $1$-shot and $5$-shot tasks, respectively.
We report the mean accuracy of randomly sampled $10k$ tasks as well as the $95\%$ confidence intervals on the testing set as mentioned in~\cite{ye2020fewshot,Zhang_2020_CVPR}.
To verify the effectiveness of our proposed DFR framework, we combined DFR with three few-shot algorithms: a commonly used baseline ProtoNet~\cite{snell2017prototypical}, two state-of-the-art methods DeepEMD~\cite{Zhang_2020_CVPR} and FEAT~\cite{ye2020fewshot}.\footnote{\ch{We utilized the official codes released by the authors, for implementations of ProtoNet, DeepEMD and FEAT and the corresponding DFR models. The results are all obtained by following the unified setting for fair comparison, which may not exactly match with the results reported in their original papers.}}
Note that we only adopt the FCN version of DeepEMD for comparison over all datasets.

\subsection{General Few-shot Classification}

We first conduct experiments on two general few-shot benchmarks: mini-ImageNet and tiered-ImageNet.

\noindent\textbf{Mini-ImageNet.}
Mini-ImageNet~\cite{vinyals2016matching} is a subset of the ILSVRC-12 challenge~\cite{krizhevsky2012imagenet} proposed for few-shot classification.
It contains 100 diverse classes with 600 images of size $84\times84\times3$ in each category.
Following the class split setting~\cite{ravi2016optimization} used in previous works, all 100 classes are divided into 64, 16 and 20 classes for training, validation and testing, respectively.

\noindent\textbf{Tiered-ImageNet.}
Similar to mini-ImageNet, Tiered-ImageNet~\cite{ren2018meta} is also a subset of ILSVRC-12, which contains more classes that are organized in a hierarchical structure, i.e., 608 classes from 34 top categories.
We follow the setups proposed by~\cite{ren2018meta}, and split 608 categories into 351, 97 and 160 for training, validation and testing, respectively.

The classification results are shown in Table~\ref{res1}.
It is clear that FEAT+DFR achieves state-of-the-art results on mini-ImageNet benchmark, while DeepEMD+DFR achieves state-of-the-art results on tiered-ImageNet benchmark.
Moreover, we observe that the improvements by DFR remain inconsistent for all baselines.
By adopting the DFR framework, the 5-way 1-shot accuracies by ProtoNet are increased by $3.0\%$ and $3.4\%$ on mini-ImageNet and tiered-ImageNet, respectively, which are even comparable to more sophisticated methods.
For the other two methods DeepEMD and FEAT, which are the current state-of-the-art FS methods, their FS classification results can still be further boosted by $1\%$ in average, after applying the DFR framework.

\begin{table}[!t]
\centering
\setlength{\tabcolsep}{0.8em}
\small
  \begin{tabular}{l|cc|cc}
     \toprule
      \multirow{2}{1cm}{ \shortstack{\textbf{Data Split}}}
     & \multicolumn{2}{c}{Class} &  \multicolumn{2}{c}{Domain}\\
     \cmidrule(r){2-3} \cmidrule(r){4-5}
      &  Train & Test  &  Source & Target\\
     \midrule
     Classic DG Setting & $-$ & $-$ & $\triangle$ & $\diamondsuit$\\
     \midrule
     Classic FS Setting & $\triangle$ & $\mathcal{S}, \mathcal{Q}$  & $-$ & $-$\\
     \midrule
     FS-DG Setting A & $\triangle$ & $\mathcal{S}, \mathcal{Q}$ & $\triangle$ & $\mathcal{S}, \mathcal{Q}$\\
     \midrule
     FS-DG Setting B & $\triangle$ & $\mathcal{S}, \mathcal{Q}$ & $\triangle$ $\mathcal{S}$ & $\mathcal{Q}$ \\
     \bottomrule
  \end{tabular}
    \caption{Comparison of different settings for DG and FS tasks. $\triangle$: training data selection for DG and FS tasks. $\diamondsuit$: testing data selection for DG tasks. $\mathcal{S}, \mathcal{Q}$: FS support and query data selection.
    The general FS and DG tasks do not split the domain and class sets, respectively.}
     \label{setting}
     \vspace{-0.2cm}
 \end{table}

\subsection{Few-shot Domain Generalization}

Domain generalization (DG) aims to learn a domain-agnostic model from multiple sources that can classify data from any target domain.
DG tasks become more challenging when there exists a class gap (besides domain gap) between the training and testing sets, i.e., DG under the few-shot setting.
General few-shot learning does not consider the influence caused by the domain gap, thus the DG models can hardly be generalized to unseen domains.
In this work, we consider a more challenging \textbf{Few-Shot Domain Generalization (FS-DG)} problem, i.e., both domain and class gaps exist between the training (source) and testing (target) sets.
In our FS-DG experiments (under both the Setting A and B), only the training samples from the source domains are selected in training.
Specifically, an $N$-way $K$-shot FS-DG task contains support and query samples from $N$ classes on the source domain in the meta-training step, and then the trained model is to predict the query data label out of the testing classes on the target domain.
Here we propose two FS-DG evaluation settings based on \textbf{different domains of support set $\mathcal{S}$} as:
(1) Setting A: Support set is only from the target domain and (2) Setting B: Support set is only from the source domain.
Both settings can evaluate the generalizability of the model, i.e., ability to extract domain-invariant and class-specific features.
Recent works~\cite{ye2020fewshot,du2021metanorm} also attempted simple FS-DG tasks to evaluate their proposed FS models.
However, only preliminary results are reported following the simple setting (i.e., Setting B in Table~\ref{setting}) without comprehensive investigation of the effect of domain gap on novel classes (test class set).
We further conduct experiments with full evaluation settings to validate the proposed DFR for FS-DG tasks using a novel FS-DomainNet Benchmark.


\subsubsection{FS-DomainNet Benchmark.}
We propose FS-DomainNet for benchmarking few-shot domain generalization.
\ch{Different from the few-shot domainnet~\cite{du2021metanorm} which only contains 200 classes with 1000 images each class, FS-DomainNet captures a much larger subset of DomainNet~\cite{peng2019moment}, i.e., 569010 images from six distinct domains (i.e., Sketch, Quickdraw, Real, Painting, Clipart and Infograph) with 345 different categories of objects from 24 divisions.
We reorganize it for few-shot learning and select all categories (i.e., 527156 images of 299 classes) that include at least the number of samples (i.e., 20) required by the $5$-shot setting on each domain. 
}
Then we split 299 categories into 191, 47 and 61 for training, validation and testing, respectively, while maintaining the consistency of class split on each domain.
More detailed descriptions and data examples of FS-DomainNet are included in our Supplementary Materials.
\ch{Different from existing few-shot benchmarks, FS-DomainNet additionally includes objects that are collected from multiple domains considering both the domain and class gaps, and the sample size varies greatly between different categories to enable more challenging FS-DG task settings.
Additionally, FS-DomainNet can also be utilized for the few-shot domain adaptation and general few-shot classification tasks.
}



\begin{table}[!t]
   \centering
   \setlength{\tabcolsep}{0.8em}
  \small 
   \begin{tabular}{lcccc}
     \toprule
     \multirow{2}{1cm}{ \textbf{Method}}  
     &  \multicolumn{2}{c}{Setting A}  & \multicolumn{2}{c}{Setting B}  \\
     \cmidrule(r){2-3} \cmidrule(r){4-5}
       &  1-shot & 5-shot & 1-shot & 5-shot \\
     \midrule
     \bf{MatchNet} & 45.23 & 54.92 & 40.61 & 49.09\\
     \midrule
     \bf{ProtoNet} & 47.96 & 66.64 & 48.70 & 67.96\\
     
     \rowcolor{Gray} \bf{ProtoNet+DFR}    & \bf{49.29} & \bf{68.73}  & \bf{49.76} & \bf{70.34}\\
     \midrule
     \bf{DeepEMD} & 53.20 & 70.59 & 51.97 & 70.62\\
     \rowcolor{Gray} \bf{DeepEMD+DFR}    & \bf{54.47} & \bf{71.60}  & \bf{54.06}  & \bf{72.33}\\
     \midrule
     \bf{FEAT}    & 51.83 & 69.26 & 52.46 & 71.54\\
     \rowcolor{Gray} \bf{FEAT+DFR}    & \bf{52.58} & \bf{69.93} & \bf{54.75} & \bf{71.91}\\
     \bottomrule
   \end{tabular}
    \caption{FS-DG Classification accuracy ($\%$) averaged on FS-DomainNet with two evaluation settings under the 5-way setting.}
     \label{fs-domainnet}
     \vspace{-0.2cm}
 \end{table}
 

\subsubsection{Experimental Setups.}
Following the classic DG setting, we choose five out of six domains from FS-DomainNet as the source domains and the remaining one as the target domain.  
We report the average FS-DG accuracies over the splits with each of the six domains as the target domain.

For $1$-shot tasks, we randomly select one support sample only from one random domain for each class;
For $5$-shot tasks, we select one labeled sample of each source domain for each class, i.e., each meta-task contains the same support samples of each domain. 
For query samples of each class with multi-domains under both $1$- and $5$-shot settings, we select the same number of query samples from each domain, i.e., $\|Q\|=3\times5=15$.

\subsubsection{Results.}
Table~\ref{fs-domainnet} shows the average accuracy of six target domains on the FS-DomainNet benchmark for two evaluation settings.
It is clear that DFR can provide consistent improvement on classification accuracies for all FS baselines under both settings.
Besides, DFR provides more significant boosting for FS-DG performance under the Setting B, thanks to its effective disentanglement of class-specific features.
Comparing to 5-shot tests, DFR provides less help for 1-shot tests, as learning from only one support sample from a random source domain for each category in meta-task is more challenging.

It is worth noting that both ProtoNet and FEAT perform better under the Setting B, while DeepEMD generates better results under the Setting A, comparing to the other setting.
It is due to the unique design of DeepEMD by adapting the channel-wise EMD metric based on the feature maps, which inexplicitly incorporates the similarity of domain information.
In the FS-DG setting A, the support and query data are from the same domain, which is, in fact, advantageous for DeepEMD, while the domain gap between support and query sets, on the contrary, degrades the DeepEMD performance under the Setting B. 
After applying the proposed DFR, the feature map in the classification branch removes the interference information, which can always improve DeepEMD under both settings.
More experiment results and analysis on the FS-DomainNet dataset can be found in our Supplementary Materials.

\begin{table}[!t]
   \centering
    \small
   \setlength{\tabcolsep}{0.4em}
  \resizebox{1.0\columnwidth}{!}{
   \begin{tabular}{lll}
     \toprule
     \multirow{2}{1cm}{ \textbf{Method}}  &     \multicolumn{2}{c}{CUB}  \\
     \cmidrule(r){2-3} 
         &  5-way 1-shot & 5-way 5-shot  \\
     \midrule
     \bf{RelationNet}~\cite{sung2018learning}   & 66.20 $\pm$ 0.99 & 82.30 $\pm$ 0.58 \\
     \bf{MAML}~\cite{finn2017model}             & 67.28 $\pm$ 1.08 & 83.47 $\pm$ 0.59 \\
     \bf{MatchNet}~\cite{vinyals2016matching}    & 71.87 $\pm$ 0.85 & 85.08 $\pm$ 0.57 \\
     \bf{COMET}~\cite{cao2021concept} & 72.20 $\pm$ 0.90 & 87.60 $\pm$ 0.50\\
     \bf{P-Transfer}~\cite{shen2021partial} & 73.88 $\pm$ 0.92 & 87.81 $\pm$ 0.48\\
     \midrule
     \bf{ProtoNet}~\cite{snell2017prototypical}  & 72.25 $\pm$ 0.21 & 87.47 $\pm$ 0.13\\
     \rowcolor{Gray} \bf{ProtoNet+DFR}    & \bf{73.52 $\pm$ 0.21} & \bf{87.90 $\pm$ 0.13} \\
     \midrule
     \bf{DeepEMD}~\cite{Zhang_2020_CVPR}    & 74.88 $\pm$ 0.30 & 88.52 $\pm$ 0.52 \\
     \rowcolor{Gray} \bf{DeepEMD+DFR}    & \bf{76.78 $\pm$ 0.29} & \bf{89.19 $\pm$ 0.52}\\
     \midrule
     \bf{FEAT}~\cite{ye2020fewshot}    & 75.68 $\pm$ 0.20 & 87.91 $\pm$ 0.13 \\
     \rowcolor{Gray} \bf{FEAT+DFR}    & \bf{77.14 $\pm$ 0.21} & \bf{88.97 $\pm$ 0.13} \\
     \bottomrule
   \end{tabular}
  }
    \caption{Fine-grained few-shot classification accuracy ($\%$) averaged on CUB with the ResNet backbone.}
     \label{res3}
     \vspace{-0.2cm}
 \end{table}


\subsection{Fine-grained Few-shot Classification}

We further evaluate DFR on a fine-grained benchmark, i.e., Caltech-UCSD Birds 200-2011 (CUB)~\cite{wah2011caltech} which was initially proposed for fine-grained image classification, which contains 200 different birds with 11788 images.
Following the split in~\cite{chen2019closer,hilliard2018few}, 200 classes are divided into 100, 50 and 50 for training, validation and testing, respectively.
We also pre-process the data by cropping each image with the provided bounding box according to the prior work~\cite{ye2020fewshot,wertheimer2021few}.

Table 4 reports the fine-grained few-shot classification results with both 5-way 1-shot and 5-way 5-shot tests.
Comparing to the general and multi-domain few-shot benchmarks that contain significant differences between the categories, fine-grained classification only includes minor intra-class differences. 
The domain information in the fine-grained dataset may contribute to the category, making it a challenging task.
It is clear that the proposed DFR can also significantly and consistently boost all FS baselines, with $0.5\%$ to $1.9\%$ additional improvements on CUB dataset.
It demonstrates that DFR can effectively remove the excursive features, and thus highlight the subtle traits which are critical for fine-grained FS classification.
\vspace{-0.2cm}
\subsection{Ablation Study}



\begin{table}[t]
\centering
\small
\begin{tabular}{cccc|c|c}
\toprule
 DFR & $\lambda_1$ & $\lambda_2$ & $\lambda_3$ & 1-shot & 5-shot\\
\midrule
 \XSolidBrush & {-} & {-} & {-} & 66.52 & 81.46 \\
 \Checkmark & 1.0 & {-} & {1.0} & 66.75 & 81.98 \\
 \Checkmark & 1.0 & 1.0 & {-} & 66.99 & 82.16 \\
 \rowcolor{Gray} \Checkmark & 1.0 & 1.0 & 1.0 & \bf{67.74} & \bf{82.49} \\
\bottomrule
\end{tabular}
\caption{Ablation study on mini-ImageNet dataset of FEAT with the proposed DFR framework.}
\label{abl}
\end{table}
We investigate the weights in our formulation and incorporate FEAT as the baseline method.
Table~\ref{abl} shows that FEAT+DFR achieves the best performance when weighting parameters are all set to $1.0$.
Compared with $L_{rec}$ and $L_{tran}$, the discriminative loss $L_{dis}$ has a more significant impact on performance as it affects the class-specific information removed from the variation branch, which is directly related to the classification ability of the classification branch.
Overall, we find that the performance is minimally affected by loss weight which also shows the robustness of our framework.


\section{Conclusion}
We propose a novel and effective Disentangled Feature Representation (DFR) framework for few-shot image classification.
Unlike the feature embeddings which may encode the excursive image information, such as background and domain, the proposed DFR aims to extract the class-specific features which is essential in most few-shot learning pipelines.
Furthermore, to tackle the challenges of the domain gap in few-shot learning, we propose a novel benchmarking dataset (FS-DomainNet) for the few-shot domain generalization task.
We have studied the importance of applying DFR in few-shot tasks by visualizing the t-SNE of the extracted features w/o DFR and disentangled features from the classification and variation branches.
Experimental results on four datasets, including three tasks (general image classification, fine-grained classification, and domain generalization) under the few-shot settings, evaluate the effectiveness of the proposed DFR framework.



\bibliography{aaai22}

\begin{thebibliography}{47}
\providecommand{\natexlab}[1]{#1}

\bibitem[{Cao, Brbic, and Leskovec(2021)}]{cao2021concept}
Cao, K.; Brbic, M.; and Leskovec, J. 2021.
\newblock Concept Learners for Few-Shot Learning.
\newblock In \emph{International Conference on Learning Representations}.

\bibitem[{Chen et~al.(2019)Chen, Liu, Kira, Wang, and Huang}]{chen2019closer}
Chen, W.-Y.; Liu, Y.-C.; Kira, Z.; Wang, Y.-C.~F.; and Huang, J.-B. 2019.
\newblock A Closer Look at Few-shot Classification.
\newblock In \emph{International Conference on Learning Representations}.

\bibitem[{Chen et~al.(2016)Chen, Duan, Houthooft, Schulman, Sutskever, and
  Abbeel}]{chen2016infogan}
Chen, X.; Duan, Y.; Houthooft, R.; Schulman, J.; Sutskever, I.; and Abbeel, P.
  2016.
\newblock InfoGAN: interpretable representation learning by information
  maximizing generative adversarial nets.
\newblock In \emph{Proceedings of the 30th International Conference on Neural
  Information Processing Systems}, 2180--2188.

\bibitem[{Du et~al.(2021)Du, Zhen, Shao, and Snoek}]{du2021metanorm}
Du, Y.; Zhen, X.; Shao, L.; and Snoek, C. G.~M. 2021.
\newblock MetaNorm: Learning to Normalize Few-Shot Batches Across Domains.
\newblock In \emph{International Conference on Learning Representations}.

\bibitem[{Fei-Fei, Fergus, and Perona(2006)}]{fei2006one}
Fei-Fei, L.; Fergus, R.; and Perona, P. 2006.
\newblock One-shot learning of object categories.
\newblock \emph{IEEE transactions on pattern analysis and machine
  intelligence}, 28(4): 594--611.

\bibitem[{Finn, Abbeel, and Levine(2017)}]{finn2017model}
Finn, C.; Abbeel, P.; and Levine, S. 2017.
\newblock Model-agnostic meta-learning for fast adaptation of deep networks.
\newblock In \emph{International Conference on Machine Learning}, 1126--1135.
  PMLR.

\bibitem[{Gidaris and Komodakis(2019)}]{gidaris2019generating}
Gidaris, S.; and Komodakis, N. 2019.
\newblock Generating classification weights with gnn denoising autoencoders for
  few-shot learning.
\newblock In \emph{Proceedings of the IEEE/CVF Conference on Computer Vision
  and Pattern Recognition}, 21--30.

\bibitem[{Hariharan and Girshick(2017)}]{hariharan2017low}
Hariharan, B.; and Girshick, R. 2017.
\newblock Low-shot visual recognition by shrinking and hallucinating features.
\newblock In \emph{Proceedings of the IEEE International Conference on Computer
  Vision}, 3018--3027.

\bibitem[{Hilliard et~al.(2018)Hilliard, Phillips, Howland, Yankov, Corley, and
  Hodas}]{hilliard2018few}
Hilliard, N.; Phillips, L.; Howland, S.; Yankov, A.; Corley, C.~D.; and Hodas,
  N.~O. 2018.
\newblock Few-shot learning with metric-agnostic conditional embeddings.
\newblock \emph{arXiv preprint arXiv:1802.04376}.

\bibitem[{Hou et~al.(2019)Hou, Chang, Ma, Shan, and Chen}]{hou2019cross}
Hou, R.; Chang, H.; Ma, B.; Shan, S.; and Chen, X. 2019.
\newblock Cross attention network for few-shot classification.
\newblock In \emph{Proceedings of the 33rd International Conference on Neural
  Information Processing Systems}, 4003--4014.

\bibitem[{Hsieh et~al.(2018)Hsieh, Liu, Huang, Fei-Fei, and
  Niebles}]{hsieh2018learning}
Hsieh, J.-T.; Liu, B.; Huang, D.-A.; Fei-Fei, L.; and Niebles, J.~C. 2018.
\newblock Learning to decompose and disentangle representations for video
  prediction.
\newblock In \emph{Proceedings of the 32nd International Conference on Neural
  Information Processing Systems}, 515--524.

\bibitem[{Jaderberg et~al.(2015)Jaderberg, Simonyan, Zisserman, and
  Kavukcuoglu}]{jaderberg2015spatial}
Jaderberg, M.; Simonyan, K.; Zisserman, A.; and Kavukcuoglu, K. 2015.
\newblock Spatial transformer networks.
\newblock In \emph{Proceedings of the 28th International Conference on Neural
  Information Processing Systems-Volume 2}, 2017--2025.

\bibitem[{Johnson, Alahi, and Fei-Fei(2016)}]{johnson2016perceptual}
Johnson, J.; Alahi, A.; and Fei-Fei, L. 2016.
\newblock Perceptual losses for real-time style transfer and super-resolution.
\newblock In \emph{European conference on computer vision}, 694--711. Springer.

\bibitem[{Krizhevsky, Sutskever, and Hinton(2012)}]{krizhevsky2012imagenet}
Krizhevsky, A.; Sutskever, I.; and Hinton, G.~E. 2012.
\newblock Imagenet classification with deep convolutional neural networks.
\newblock In \emph{Advances in neural information processing systems},
  1097--1105.

\bibitem[{Lee et~al.(2020)Lee, Tseng, Mao, Huang, Lu, Singh, and
  Yang}]{lee2020drit++}
Lee, H.-Y.; Tseng, H.-Y.; Mao, Q.; Huang, J.-B.; Lu, Y.-D.; Singh, M.; and
  Yang, M.-H. 2020.
\newblock Drit++: Diverse image-to-image translation via disentangled
  representations.
\newblock \emph{International Journal of Computer Vision}, 128(10): 2402--2417.

\bibitem[{Lee et~al.(2019)Lee, Maji, Ravichandran, and Soatto}]{lee2019meta}
Lee, K.; Maji, S.; Ravichandran, A.; and Soatto, S. 2019.
\newblock Meta-learning with differentiable convex optimization.
\newblock In \emph{Proceedings of the IEEE/CVF Conference on Computer Vision
  and Pattern Recognition}, 10657--10665.

\bibitem[{Li et~al.(2019)Li, Eigen, Dodge, Zeiler, and Wang}]{li2019finding}
Li, H.; Eigen, D.; Dodge, S.; Zeiler, M.; and Wang, X. 2019.
\newblock Finding task-relevant features for few-shot learning by category
  traversal.
\newblock In \emph{Proceedings of the IEEE/CVF Conference on Computer Vision
  and Pattern Recognition}, 1--10.

\bibitem[{Li et~al.(2020{\natexlab{a}})Li, Zhang, Li, and
  Fu}]{li2020adversarial}
Li, K.; Zhang, Y.; Li, K.; and Fu, Y. 2020{\natexlab{a}}.
\newblock Adversarial feature hallucination networks for few-shot learning.
\newblock In \emph{Proceedings of the IEEE/CVF Conference on Computer Vision
  and Pattern Recognition}, 13470--13479.

\bibitem[{Li et~al.(2020{\natexlab{b}})Li, Jin, Lin, Liu, Wu, Yu, Zhou, and
  Chen}]{li2020learning}
Li, X.; Jin, X.; Lin, J.; Liu, S.; Wu, Y.; Yu, T.; Zhou, W.; and Chen, Z.
  2020{\natexlab{b}}.
\newblock Learning Disentangled Feature Representation for Hybrid-distorted
  Image Restoration.
\newblock In \emph{European Conference on Computer Vision}, 313--329. Springer.

\bibitem[{Li et~al.(2021)Li, Xu, Wei, and Deng}]{li2021generalized}
Li, X.; Xu, Z.; Wei, K.; and Deng, C. 2021.
\newblock Generalized Zero-Shot Learning via Disentangled Representation.
\newblock In \emph{Proceedings of the AAAI Conference on Artificial
  Intelligence}, volume~35, 1966--1974.

\bibitem[{Liu et~al.(2021)Liu, Fu, Xu, Yang, Li, Wang, and
  Zhang}]{liu2021learning}
Liu, C.; Fu, Y.; Xu, C.; Yang, S.; Li, J.; Wang, C.; and Zhang, L. 2021.
\newblock Learning a Few-shot Embedding Model with Contrastive Learning.
\newblock In \emph{Proceedings of the AAAI Conference on Artificial
  Intelligence}, volume~35, 8635--8643.

\bibitem[{Liu et~al.(2019)Liu, Huang, Mallya, Karras, Aila, Lehtinen, and
  Kautz}]{liu2019few}
Liu, M.-Y.; Huang, X.; Mallya, A.; Karras, T.; Aila, T.; Lehtinen, J.; and
  Kautz, J. 2019.
\newblock Few-shot unsupervised image-to-image translation.
\newblock In \emph{Proceedings of the IEEE/CVF International Conference on
  Computer Vision}, 10551--10560.

\bibitem[{Liu, Schiele, and Sun(2020)}]{Liu2020E3BM}
Liu, Y.; Schiele, B.; and Sun, Q. 2020.
\newblock An Ensemble of Epoch-wise Empirical Bayes for Few-shot Learning.
\newblock In \emph{European Conference on Computer Vision (ECCV)}.

\bibitem[{Nichol, Achiam, and Schulman(2018)}]{nichol2018first}
Nichol, A.; Achiam, J.; and Schulman, J. 2018.
\newblock On first-order meta-learning algorithms.
\newblock \emph{arXiv preprint arXiv:1803.02999}.

\bibitem[{Oreshkin, L{\'o}pez, and Lacoste(2018)}]{oreshkin2018tadam}
Oreshkin, B.~N.; L{\'o}pez, P.~R.; and Lacoste, A. 2018.
\newblock TADAM: Task dependent adaptive metric for improved few-shot learning.
\newblock In \emph{NeurIPS}.

\bibitem[{Peng et~al.(2019)Peng, Bai, Xia, Huang, Saenko, and
  Wang}]{peng2019moment}
Peng, X.; Bai, Q.; Xia, X.; Huang, Z.; Saenko, K.; and Wang, B. 2019.
\newblock Moment matching for multi-source domain adaptation.
\newblock In \emph{Proceedings of the IEEE/CVF International Conference on
  Computer Vision}, 1406--1415.

\bibitem[{Prabhudesai et~al.(2021)Prabhudesai, Lal, Patil, Tung, Harley, and
  Fragkiadaki}]{prabhudesai2021disentangling}
Prabhudesai, M.; Lal, S.; Patil, D.; Tung, H.-Y.; Harley, A.~W.; and
  Fragkiadaki, K. 2021.
\newblock Disentangling 3D Prototypical Networks for Few-Shot Concept Learning.
\newblock In \emph{International Conference on Learning Representations}.

\bibitem[{Ravi and Larochelle(2017)}]{ravi2016optimization}
Ravi, S.; and Larochelle, H. 2017.
\newblock Optimization as a model for few-shot learning.
\newblock In \emph{International Conference on Learning Representations}.

\bibitem[{Ren et~al.(2018)Ren, Triantafillou, Ravi, Snell, Swersky, Tenenbaum,
  Larochelle, and Zemel}]{ren2018meta}
Ren, M.; Triantafillou, E.; Ravi, S.; Snell, J.; Swersky, K.; Tenenbaum, J.~B.;
  Larochelle, H.; and Zemel, R.~S. 2018.
\newblock Meta-learning for semi-supervised few-shot classification.
\newblock In \emph{ICLR}.

\bibitem[{Rusu et~al.(2019)Rusu, Rao, Sygnowski, Vinyals, Pascanu, Osindero,
  and Hadsell}]{rusu2018meta}
Rusu, A.~A.; Rao, D.; Sygnowski, J.; Vinyals, O.; Pascanu, R.; Osindero, S.;
  and Hadsell, R. 2019.
\newblock Meta-Learning with Latent Embedding Optimization.
\newblock In \emph{International Conference on Learning Representations}.

\bibitem[{Shen et~al.(2021)Shen, Liu, Qin, Savvides, and
  Cheng}]{shen2021partial}
Shen, Z.; Liu, Z.; Qin, J.; Savvides, M.; and Cheng, K.-T. 2021.
\newblock Partial Is Better Than All: Revisiting Fine-tuning Strategy for
  Few-shot Learning.
\newblock In \emph{Proceedings of the AAAI Conference on Artificial
  Intelligence}, volume~35, 9594--9602.

\bibitem[{Simon et~al.(2020)Simon, Koniusz, Nock, and
  Harandi}]{simon2020adaptive}
Simon, C.; Koniusz, P.; Nock, R.; and Harandi, M. 2020.
\newblock Adaptive subspaces for few-shot learning.
\newblock In \emph{Proceedings of the IEEE/CVF Conference on Computer Vision
  and Pattern Recognition}, 4136--4145.

\bibitem[{Snell, Swersky, and Zemel(2017)}]{snell2017prototypical}
Snell, J.; Swersky, K.; and Zemel, R. 2017.
\newblock Prototypical networks for few-shot learning.
\newblock In \emph{Proceedings of the 31st International Conference on Neural
  Information Processing Systems}, 4080--4090.

\bibitem[{Sung et~al.(2018)Sung, Yang, Zhang, Xiang, Torr, and
  Hospedales}]{sung2018learning}
Sung, F.; Yang, Y.; Zhang, L.; Xiang, T.; Torr, P.~H.; and Hospedales, T.~M.
  2018.
\newblock Learning to Compare: Relation Network for Few-Shot Learning.
\newblock In \emph{Proceedings of the IEEE Conference on Computer Vision and
  Pattern Recognition}.

\bibitem[{Tang, Wertheimer, and Hariharan(2020)}]{tang2020revisiting}
Tang, L.; Wertheimer, D.; and Hariharan, B. 2020.
\newblock Revisiting pose-normalization for fine-grained few-shot recognition.
\newblock In \emph{Proceedings of the IEEE/CVF Conference on Computer Vision
  and Pattern Recognition}, 14352--14361.

\bibitem[{Tian et~al.(2020)Tian, Wang, Krishnan, Tenenbaum, and
  Isola}]{tian2020rethinking}
Tian, Y.; Wang, Y.; Krishnan, D.; Tenenbaum, J.~B.; and Isola, P. 2020.
\newblock Rethinking few-shot image classification: a good embedding is all you
  need?
\newblock In \emph{Computer Vision--ECCV 2020: 16th European Conference,
  Glasgow, UK, August 23--28, 2020, Proceedings, Part XIV 16}, 266--282.
  Springer.

\bibitem[{Tokmakov, Wang, and Hebert(2019)}]{tokmakov2019learning}
Tokmakov, P.; Wang, Y.-X.; and Hebert, M. 2019.
\newblock Learning compositional representations for few-shot recognition.
\newblock In \emph{Proceedings of the IEEE/CVF International Conference on
  Computer Vision}, 6372--6381.

\bibitem[{Van~der Maaten and Hinton(2008)}]{van2008visualizing}
Van~der Maaten, L.; and Hinton, G. 2008.
\newblock Visualizing data using t-SNE.
\newblock \emph{Journal of machine learning research}, 9(11).

\bibitem[{Vinyals et~al.(2016)Vinyals, Blundell, Lillicrap, Kavukcuoglu, and
  Wierstra}]{vinyals2016matching}
Vinyals, O.; Blundell, C.; Lillicrap, T.; Kavukcuoglu, K.; and Wierstra, D.
  2016.
\newblock Matching networks for one shot learning.
\newblock In \emph{Proceedings of the 30th International Conference on Neural
  Information Processing Systems}, 3637--3645.

\bibitem[{Wah et~al.(2011)Wah, Branson, Welinder, Perona, and
  Belongie}]{wah2011caltech}
Wah, C.; Branson, S.; Welinder, P.; Perona, P.; and Belongie, S. 2011.
\newblock The caltech-ucsd birds-200-2011 dataset.

\bibitem[{Wang et~al.(2018)Wang, Girshick, Hebert, and Hariharan}]{wang2018low}
Wang, Y.-X.; Girshick, R.; Hebert, M.; and Hariharan, B. 2018.
\newblock Low-shot learning from imaginary data.
\newblock In \emph{Proceedings of the IEEE conference on computer vision and
  pattern recognition}, 7278--7286.

\bibitem[{Wertheimer, Tang, and Hariharan(2021)}]{wertheimer2021few}
Wertheimer, D.; Tang, L.; and Hariharan, B. 2021.
\newblock Few-Shot Classification With Feature Map Reconstruction Networks.
\newblock In \emph{Proceedings of the IEEE/CVF Conference on Computer Vision
  and Pattern Recognition}, 8012--8021.

\bibitem[{Xu et~al.(2020)Xu, Wang, Tu et~al.}]{xu2020attentional}
Xu, W.; Wang, H.; Tu, Z.; et~al. 2020.
\newblock Attentional Constellation Nets for Few-Shot Learning.
\newblock In \emph{International Conference on Learning Representations}.

\bibitem[{Yang, Liu, and Xu(2021)}]{yang2021free}
Yang, S.; Liu, L.; and Xu, M. 2021.
\newblock Free Lunch for Few-shot Learning: Distribution Calibration.
\newblock In \emph{International Conference on Learning Representations}.

\bibitem[{Ye et~al.(2020)Ye, Hu, Zhan, and Sha}]{ye2020fewshot}
Ye, H.-J.; Hu, H.; Zhan, D.-C.; and Sha, F. 2020.
\newblock Few-Shot Learning via Embedding Adaptation with Set-to-Set Functions.
\newblock In \emph{IEEE/CVF Conference on Computer Vision and Pattern
  Recognition (CVPR)}, 8808--8817.

\bibitem[{Zhang et~al.(2020)Zhang, Cai, Lin, and Shen}]{Zhang_2020_CVPR}
Zhang, C.; Cai, Y.; Lin, G.; and Shen, C. 2020.
\newblock DeepEMD: Few-Shot Image Classification With Differentiable Earth
  Mover's Distance and Structured Classifiers.
\newblock In \emph{IEEE/CVF Conference on Computer Vision and Pattern
  Recognition (CVPR)}.

\bibitem[{Zhao et~al.(2021)Zhao, Yang, Lin, Yang, and He}]{zhao2021looking}
Zhao, J.; Yang, Y.; Lin, X.; Yang, J.; and He, L. 2021.
\newblock Looking Wider for Better Adaptive Representation in Few-Shot
  Learning.
\newblock In \emph{Proceedings of the AAAI Conference on Artificial
  Intelligence}, volume~35, 10981--10989.

\end{thebibliography}




\end{document}